Article

# Architecture for Trajectory-Based Fishing Ship Classification with AIS Data

David Sánchez Pedroche, Daniel Amigo, Jesús García and José Manuel Molina

# Architecture for Trajectory-Based Fishing Ship Classification with AIS Data

**David Sánchez Pedroche *, Daniel Amigo *, Jesús García and José Manuel Molina**

Group GIAA, University Carlos III of Madrid, 28270 Madrid, Spain; jgherrer@inf.uc3m.es (J.G.); molina@ia.uc3m.es (J.M.M.)

* Correspondence: davsanch@inf.uc3m.es (D.S.P.); damigo@inf.uc3m.es (D.A.)

**Abstract:** This paper proposes a data preparation process for managing real-world kinematic data and detecting fishing vessels. The solution is a binary classification that classifies ship trajectories into either fishing or non-fishing ships. The data used are characterized by the typical problems found in classic data mining applications using real-world data, such as noise and inconsistencies. The two classes are also clearly unbalanced in the data, a problem which is addressed using algorithms that resample the instances. For classification, a series of features are extracted from spatiotemporal data that represent the trajectories of the ships, available from sequences of Automatic Identification System (AIS) reports. These features are proposed for the modelling of ship behavior but, because they do not contain context-related information, the classification can be applied in other scenarios. Experimentation shows that the proposed data preparation process is useful for the presented classification problem. In addition, positive results are obtained using minimal information.

## 1. Introduction

Maritime vigilance is essential to ensure safety and security at sea. Illegal, unreported and unregulated (IUU) fishing [1] poses a risk to food safety and maritime biodiversity. It has been estimated that between 11 and 26 million tons of fish are caught annually by these illegal fishing activities, accounting for approximately 15% of fish consumed globally [2]. To counter this illegal activity, maritime vigilance systems need the capability to locate vessels inside an area and recognize fishing ships. Multiple sensors are capable of providing the kinematic information of located objects to address this localization problem. These sensors can be categorized into two major groups: those relying on the collaboration of the object using information that is provided by the located ship (e.g., Automatic Identification System, AIS), and those that only use the information generated by the sensor (e.g., primary radar). Non-collaborative sensors only provide kinematic information without ship characteristics, while collaborative sensors, such as the AIS, provide additional information that could assist ship identification. This additional information, however, is susceptible to manipulation by the ship owner, resulting in a loss of trust.

This paper proposes ship type identification using only the kinematic information that can be extracted from any sensor. Furthermore, the proposed process allows classification of objects as either fishing ships or other types of vessels.

A significant consideration in location and recognition problems is the context of operation and the usable information related to that context. A simple example is information about the specific areas in which fishing boats may be commonly found or, in contrast, areas in which fishing ships are rare. This context information can be a useful tool for an identification system because it can assist identification, e.g., it is highly likely that a ship detected in a fishing area is a fishing ship. However, this improvement in recognition capability also specializes the system to the environment of the particular context information. As a result, if the context information differs between environments, it is necessary to develop a different system for each environment.

In addition, the specialization of the system implies that classification of trajectories heavily relies on the context, to the extent that the context is considered to be more important than ship behavior. Thus, if a classifier is able to locate all fishing ships inside a specific area, an illegal fishing ship could easily cheat the system by not entering the modelled area and fishing illegally elsewhere.

As a result of these issues, the approach proposed in this paper does not consider context information and detects fishing ships using only sensor-extracted kinematics. Notably, the kinematic data used in the proposed approach does not identify the position of the ship, because this measure would limit the system to a particular scenario. Thus, using exhaustive data processing, it is possible to train the classification algorithm to perform ship type detection (i.e., fishing in this paper) with minimal information; that is, the aim is to achieve classification using the minimum amount of inputs.

The original data used for the process were real-world data without any treatment. In any data mining problem, problems inherent to the raw data exist, such as [3]:

- Data instances exist that do not provide information to achieve the desired task, presenting inconsistencies, null values, or extreme values.
- Noisy sensor measurements, which may ruin the results.
- The number of objects (e.g., ships) of each type is not equivalent and, therefore, there is an imbalance in the data.
- Movements made by different objects (e.g., ships) may differ greatly in terms of both distance and duration, which implies that they are not comparable to each other.

To solve these problems, the proposed process takes the following steps: First, data cleaning is undertaken to eliminate information that is inconsistent or incorrect, or otherwise cannot be used in the subsequent classification process. Second, a state estimation filter is used to smooth the trajectory of different ships, thus reducing the influence of atypical measures and noisy data. Before proceeding with classification, since there are clear differences in the sizes of trajectories, it is necessary to apply segmentation to allow entries to the classification problem to be compared. Finally, once the preprocessed data are obtained, a class balancing step is necessary due to the significant difference in the number of ships of different types.

Following data preparation, the inputs and outputs to the classification algorithm must be defined before the algorithm can be applied. The process output is the defined class (i.e., a fishing ship or a non-fishing ship). However, different track measurements cannot be used simply as the inputs; measurements must be translated into the track kinematic values to represent the ship behavior.

The approach proposed in this paper is the systematization of a process that uses different sub-processes with the objective of achieving a better classification. Thus, the method observes the relationships of the different sub-processes and analyzes their impact within the final classification. The results are analyzed from a multi-objective perspective to consider both the success of the classification in a generalist way, and the resolution of the problem taking into account the existing imbalance.

The results obtained show that the proposed data preparation process is successful, resulting in improved classification when the steps of the process are applied. In addition, classification is achieved using minimal kinematic information.

This paper represents an application of the proposal of [4,5], in which the trajectory classification problem was considered using multiple inputs with a lower level of preprocessing, and the inputs were analyzed to identify those useful for classification.

This paper is organized as follows: In Section 2 work related to this problem is discussed. In Section 3 the proposed process is explained, and in Section 4 the experiments are outlined and results are shown. Finally, the conclusions and perspectives for future research are presented in Section 5.

## 2. Related Work

In [4,5] a preliminary study was made of the problem presented in this paper. This study considered the problem holistically, supporting the non-binary classification of ships, rather than considering only the detection of a particular type. The main difference from the current proposal is the special focus on preprocessing to prepare the data.

In the current study, preprocessing was undertaken for the treatment of erroneous variables and to adapt the approach to the detection of fishing ships. The cleaning undertaken in the current approach was largely possible due to previous research, based on which data errors were detected, prior to being corrected during the preprocessing step of this proposal.

Previous studies explored a broader set of variables and differed in their behavioral approach compared to the current paper. In this study, a number of previously studied variables were discarded, leaving only those that provided better results, and considering that only non-context information is desired in the proposed approach.

Previous research also provided information about the configuration of relevant algorithms, such as the interacting multiple model (IMM) filter, thus allowing the optimal methodology to be chosen for the current method. In addition, to address issues identified in previous research, particular consideration is given in the proposed approach to the problem of balancing.

In the development of the approach proposed in this paper, analysis was undertaken of previous studies of data cleaning and filtering approaches to reduce atypical measurements and smooth trajectories, the treatment of unbalanced data, and the trajectory classification problem, considering the different classification algorithms and the feature extraction of ship tracks to represent each ship's behavior. The following sections review the literature relevant to each of these issues, in addition to outlining the approaches used for similar problems.

### *2.1. Data Preprocessing*

Raw data must be prepared for operation within a classifier, so a cleaning process is necessary to transform the data into a useful input. Specifically, the presented approach transforms a series of dispersed measures of multiple ships into a series of ordered trajectories.

The Automatic Identification System (AIS) [6] is mandated to be fitted to most maritime vehicles by the International Maritime Organization (IMO) [7]. In this study, AIS data was chosen because it is widely used in the literature and freely available, in contrast to other sensors. AIS data can differ between the type of transponder (class A or B) and messages can contain different information, however, the data contains kinematic information (timestamp, GPS measurements in WGS-84 coordinates, an orientation in relation to north and speed) and static information about the ship, such as ship name, ship dimensions, maneuver, ship type, and Maritime Mobile Service Identity (MMSI). The possible values of ship type are shown in Table 1.

**Table 1.** Ship types provided by the Automatic Identification System (AIS).

| | | |
|---|---|---|
| Anti-pollution | Cargo | Dredging |
| Fishing | HSC | Pilot |
| Port tender | Military | Passenger |
| Law enforcement | Pleasure | Medical |
| Reserved | Sailing | SAR |
| Tanker | Towing | Tug |

Nevertheless, due to problems with this information resource [8], preprocessing is necessary to treat possible inconsistencies, null values, and noise. No algorithms exist to treat data inconsistencies and null values, thus, the most effective solution is to analyze the data and the underlying problem.

However, to address track noise generated by kinematic AIS data over time, so-called state estimation algorithms exist in the field of data fusion. Using a probability approach, these algorithms can increase confidence in the measurements and minimize the impact of noise from sensors [9]. For AIS, the state is presented in terms of latitude and longitude, measured by a GPS device inside the ship.

The Kalman filter (KF) [10] is a well-known state estimation filter, although its extended version (EKF) is more suitable to a non-linear problem, such as that proposed [11]. Unlike the KF, the EKF is not optimal; however, research into particle filters [12] or adaptive filters has aimed to solve this problem. For the process proposed in this paper, the interacting multiple model (IMM) filter [13] was selected for being a solid solution to this problem [14].

### *2.2. Data Imbalance*

Because of its impact in multiple areas and existence in reality, class imbalance is an extensively researched problem in the field of data mining. Multiple approaches have been proposed as solutions, which be divided into methods using oversampling and undersampling to resample (add or remove data-level instances) [15,16], and those that use specific algorithms that consider the imbalance in the classification process [17,18].

In addition, hybrid approaches exist that combine several of these techniques in a single solution.

Due to their proven effectivity, relatively low computational cost, and simple implementation, two of the best known and most used at the data level are:

- Random undersampling: This technique is a simple solution that reduces the instances of the majority class by randomly deleting instances [19]. Its major drawback is the possibility of excluding useful data.
- Synthetic Minority Over-sampling Technique (SMOTE): The efficiency of this widely known oversampling technique has been proven in different imbalance problems; it creates new instances at a random point along the line segment that joins two neighbors (calculated by the k-nearest neighbors algorithm) of the original instances [20].

In classification problems, the main algorithm performance evaluation metric is the success rate, also known as accuracy, which is the ratio of the correctly classified instances to the total instances of the test set. This metric is not enough to measure the results in an imbalanced dataset since it does not consider the entire problem.

The imbalance problem is demonstrated in the example confusion matrix presented in Table 2. In this example, the negative class clearly reveals significantly more instances than the positive class. This implies an 85% success rate, but also achieves a poor result in the positive class classification, as most instances are classified to the negative class.

**Table 2.** Confusion matrix example.

| | **Predicted as Positive** | **Predicted as Negative** |
|---|---|---|
| Positive Class | TP = 1234 | FN = 4989 |
| Negative Class | FP = 361 | TN = 28,730 |

From the confusion matrix, it is possible to obtain other metrics to evaluate specific parts of the classification, such as sensitivity or precision. The F-measure is a widely used metric that combines sensitivity and precision for analyzing the positive class [16,19], providing more specific information than the accuracy, but in exchange losing the overall view of the classification results. For example, in the previous confusion matrix, the F-measure is approximately 32%, which indicates poor positive class classification even though the accuracy was good. Therefore, in order to evaluate classification in imbalanced problems, it is necessary to take into consideration both accuracy and the F-measure (or a similar metric).

Class imbalance is common in datasets used in trajectory classification problems, but a small number of authors, such as [21], have used SMOTE and emphasized solving the problem using the resample algorithms that aid classification. In contrast, other papers [4,22–24] argue that the classification is valid by analyzing the results or the metrics.

### 2.3. Classification Problem

Multiple algorithms exist that enable the prediction of a class using a set of input variables. The most common groups of techniques are logic based, perceptron based, statistic learning based, and the Support Vector Machine (SVM) [3]. Many of these algorithms can generate good results. In this study, SVM and decision trees were selected, as the examined problem involves a binary classification to predict when a boat is a fishing ship or a non-fishing ship.

In addition to the classification results, the decision tree algorithm also provides an understanding of the use of the input data via analysis of the tree decision nodes. Using a predictor importance function, it is possible to estimate the relevance in the classification of each input [25].

In the literature, several approaches exist to address problems similar to those outlined in this article. These can be grouped by the source of information they use. Those that use images to detect the type of ship, including Synthetic Aperture Radar (SAR) [26,27] or photographs [28], use a different approach based on analysis of the pixels of static images. Comparative (terrain reference) navigation represents an approach similar to the use of images to explore the environment, although it does not require a satellite; in [29] the use of 3D multibeam sonar data is explored. Other approaches use sensor trajectories, such as AIS or RADAR.

The authors of [24] extract both kinematic and context-specific information from each trajectory to classify the type of ship. These geographical characteristics, such as the distance to the coast or clustering of areas in which the main ship type is defined, are useful for the classification problem. However, this approach, and that used in ref. [30], is not used in this paper. Ship characteristics extracted from the AIS data, such as dimensions, represent context information that is not specific to a given scenario, so meet one criterion for use in the proposed method. However, because this information is not available from a range of sensor types, this data source does not satisfy our requirement for a system that is usable with any sensor that provides kinematic information. These two solutions also differ in the use of full trajectories (by compressing the data prior to the classification stage), compared to the proposal of this paper in which the classification uses segments of the trajectory to introduce equivalent information fragments.

In [31], the trajectory is divided and features are extracted according to kinematics only. This approach was also used by the authors of [22] without dividing the trajectory, using a multivariate time series classifier as a specific algorithm for time series. Both of these papers were tested with a specific dataset; the first was of fishing and cargo vessels in a limited area, and the second was a small dataset with highly unbalanced classes and used a boosting classifier with the objective of maximizing the accuracy by learning the models of the classes. In [32], the authors explored the problem of classification using neural networks. However, this approach used positional data, which were not considered for the current proposal. Another similar approach was presented in [33], which used a random forest to classify different trajectories in two dimensions, instead of boat trajectories.

Another type of target was classified in [34–36], in which airplanes or land vehicles, rather than ships, were classified using characteristics extracted from trajectories.

## 3. Proposed Architecture

The proposed architecture, as shown in Figure 1, uses a AIS dataset of measurements that represent ship trajectories. The first step in the preparation of the trajectories is to clean the data by removing inconsistencies, and null and incorrect values. The second step is to use the IMM filter to smooth the trajectories, thus reducing the impact of noisy data. The segmentation step divides the trajectories into equal size segments for comparability in the classification. Prior to classification, a further step is required to address the issue of data imbalance. The features extraction step models the behavior of each ship from its trajectory kinematics before the final step, in which the classification algorithm is applied and results analyzed. The following sections summarize each of the main steps of the proposed process.

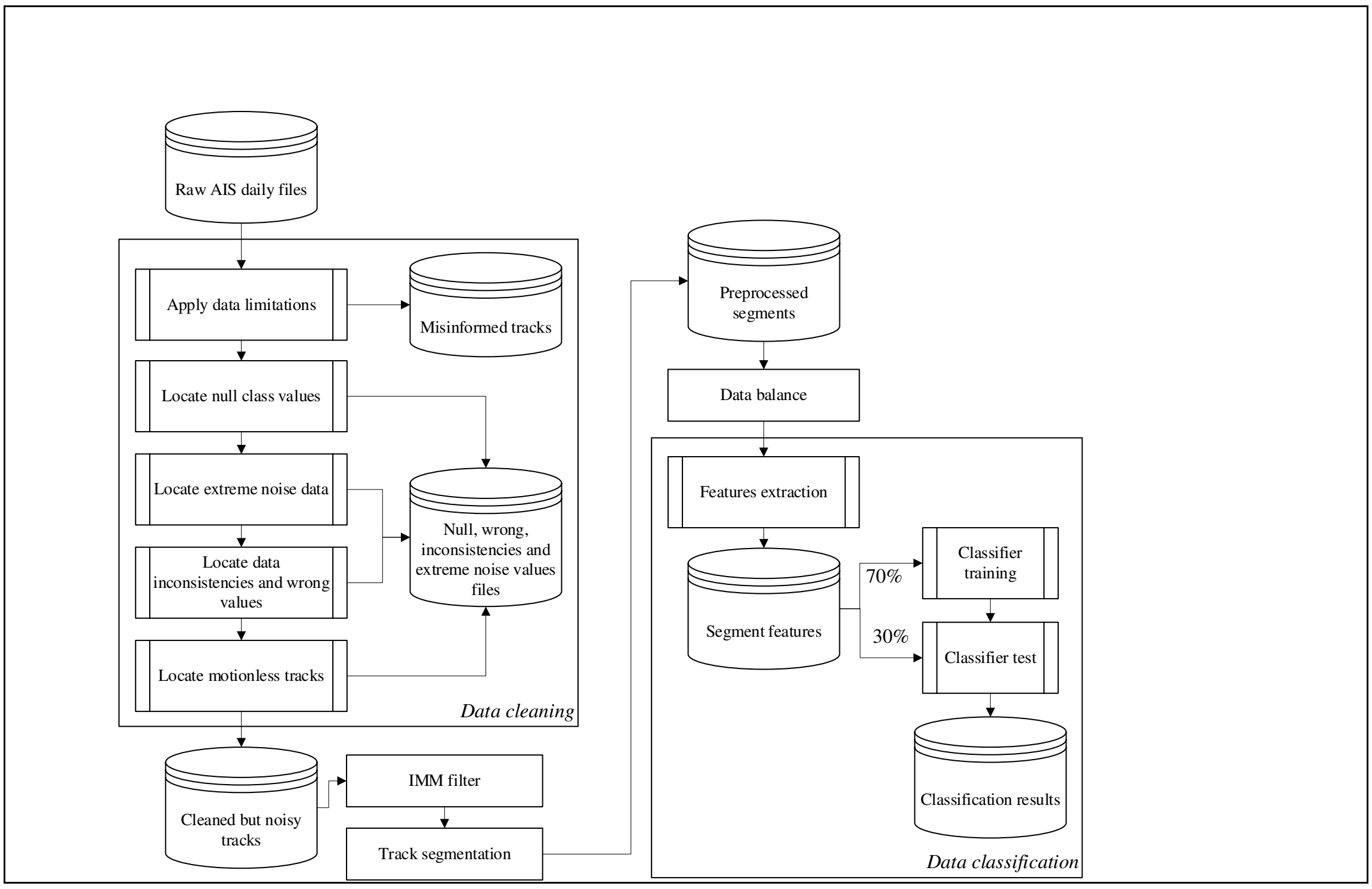

**Figure 1.** Architecture scheme. The code of the proposed architecture is publicly accessible at GitHub (https://github.com/DanielAmigo/trajectory-based-ship-classification).

### 3.1. Data Source

The data source must be a set of kinematic information extracted from the measurements of one or more sensors and with a class indicating if the instance is from a fishing ship. The chosen resource is AIS data because its ship type record can be used to identify if the measurement is provided by a fishing ship, and due to the high availability of this data source.

The selected AIS data is a repository provided by the Danish Maritime Authority [37], and is a recompilation of millions of AIS contacts off the coast of Denmark since 2006. Divided into daily files of approximately 1.8 GB, the data represents a practically unlimited amount of information, and is thus sufficient to generate an input dataset for the study objective. The data contains the Maritime Mobile Service Identity (MMSI), which makes it possible to distinguish ship trajectories, and the ship type value, which allows extraction of a class for training the classification algorithm.

### 3.2. Data Cleaning

The data cleaning process begins by dividing the data into trajectories to identify possible inconsistencies. MMSI, the ship identifier provided with the AIS data, is used to divide the trajectories. In this division, the algorithm makes an initial noise reduction based on the limitation of each trajectory. The limitations are applied over the time gap between track measurements and over the minimum trajectory size. The slowest AIS refresh rate on a moving ship is 10 s [38], thus, the time gap limitation assumes that the track measurements must be separated by no more than 11 s, allowing for a minimum delay between transmitter and receiver.

To ensure sufficient information for the classification stage, the process limits the minimum track size to 50 measurements. This limit was chosen for its best results among those tested, and is the minimum size for a trajectory belonging to a ship that contains enough information. Figure 2b shows a noisy track in which the AIS data transmission rate is not achieved, and therefore, some of its measurements cannot be used for the filtering algorithm.

After track division, the cleaning process removes inconsistencies and null or wrong values. The first step in this process is to locate high offsets in the WGS-84 coordinates of the measurements. Figure 2a shows an example of an extreme noise value that is deleted from the dataset. In addition, tracks that does not contain the ship type or those that do not correspond to a ship (the set includes measurements from ground stations) are removed from the dataset.

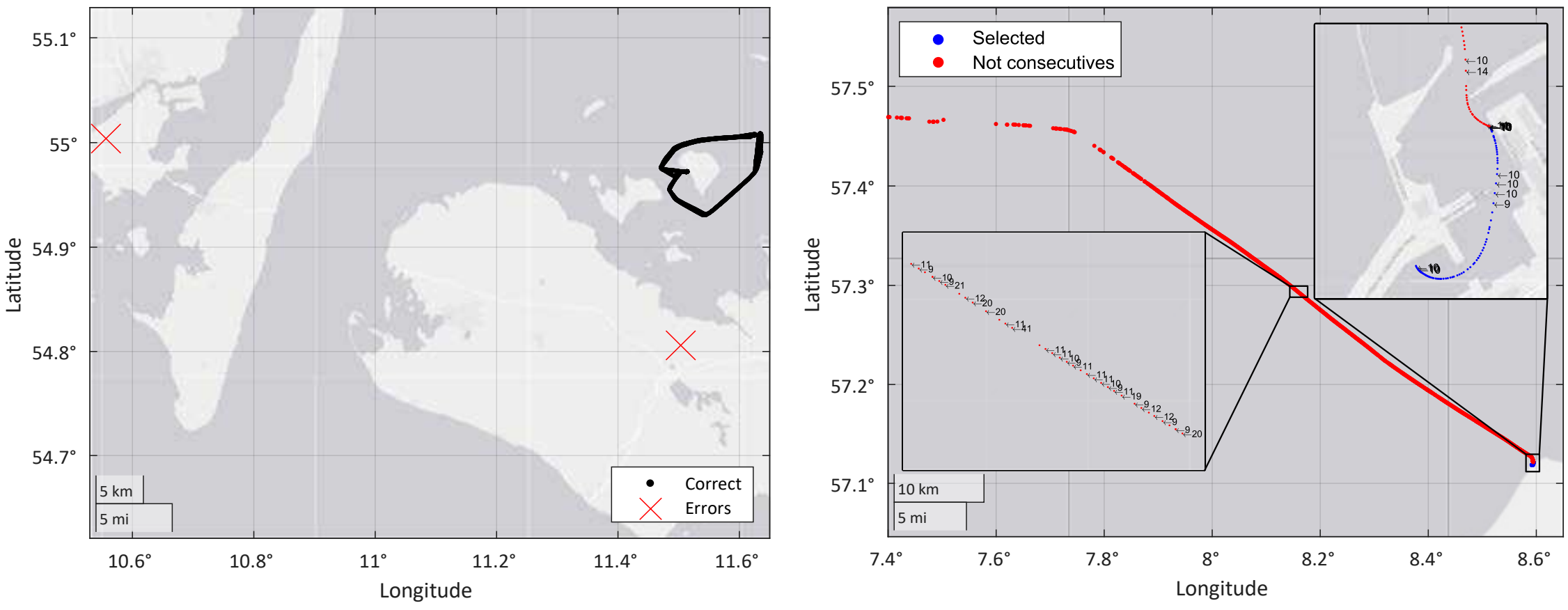

(**a**) (**b**)

**Figure 2.** (**a**) Example of extreme noise. (**b**) Example of the time gap problem.

Because the AIS data provides a maneuver value, it is possible to find inconsistencies in the data such as non-maneuvering trajectories that indicate movements within the measurements. These inconsistencies imply a wrong value in the AIS transmitter that could involve other incorrect data points, thus the process removes those inconsistences from the dataset. Trajectories that do not show motion are also removed because they are not useful for the proposed problem, which uses trajectory kinematics to perform classification.

## 3.3. Data Filtering

To reduce the impact of measurement noise, the IMM filtering algorithm is applied to smooth the trajectories, thus resulting in more robust kinematic ship information. The IMM filter can be configured by modifying the different models of prediction (linear movement, turns, acceleration, etc.) and the switch probabilities between modes. The system proposed in this paper is concerned with ship movements and therefore is modeled using two modes: the first mode models all of the linear movements, and the second models all of the noisy movements (acceleration, deceleration, or curvilinear movements); thus, the second movement mode represents all of the movements that differ from linear movement.

To implement the prediction equations of each mode, the algorithm uses an EKF, and the second mode is more sensitive to this measure. The measure error is set to 10 m, which is the average noise of a GPS sensor [39].

In the selection of the prediction mode as part of the filter configuration, it is possible to modify the switch probabilities between modes, thus making the filter more robust or sensitive to variation in linear movements. This parameter must be set as a compromise between noise robustness and the detection of maneuvers; although it is important to detect maneuvers due to ship acceleration or course variations, it is also necessary that noise measurements are not mistaken for maneuvers.

### 3.4. Trajectories Segmentation

The classification algorithm requires comparable inputs to obtain positive results, but the trajectories obtained after the preprocessing step have large size differences, meaning that the quantity of information could differ significantly, thus preventing their comparison.

To solve this problem, the proposed approach identifies cutting points in the trajectories that can be used to generate smaller fragments; that is, sub-trajectories are used to represent the main trajectory. This is achieved using a uniform sampling algorithm which makes segments of a fixed size, meaning that the sub-trajectories result from cuts made at a constant increment. In previous studies, the measure of 50 AIS plots was used for this division. Thus, this measure was also used in current study, with the aim making comparisons between the different balancing techniques evaluated in this paper.

### 3.5. Data Imbalance Treatment

The dataset contains a clear class imbalance that must be resolved prior to classification as it could distort the results. This imbalance problem arises from the difference in the numbers of each type of ship in the dataset, with cargo ships having the most instances in the dataset.

This problem is resolved using an algorithm prior to the classification stage. Experimentation was conducted to examine how the selection of this algorithm affects the classification results.

The two algorithms chosen to solve the imbalance problem are explained in Section 2: random undersampling and SMOTE. In both cases, the objective of the balancing solution is to generate a dataset in which 50% of the instances are fishing ships, and the other 50% are non-fishing ships. In the case of random undersampling, entries of non-fishing vessels are removed randomly, regardless of the ship type. In SMOTE, fishing instances are created, until a threshold of 50% of the dataset is reached.

The classification success of the solution to the data imbalance problem is measured with the accuracy metric and the F-measure with $\sigma = 1$.

### 3.6. Classification

#### 3.6.1. Feature Extraction

To classify the trajectories, features are extracted from the tracks that enable ship behavior to be modeled without considering its context of operation. The proposed approach identifies kinematic variables that represent this behavior without considering the position of the target, since it is assumed that the behavior of the ship should be the same in any operating environment. Thus, it is intended that the classification algorithm does not learn from elements specific to the situation.

The kinematic values relevant to the presented problem were chosen based on the study in [24], which presents a set of variables that model ship behavior. Specifically, variables can be identified that are related to:

- The course variation: to describe the travel direction and changes in direction.
- The distance: to characterize the trajectory's range and complexity.
- The speed: using the vector norm and the variation of each point to describe the different speed properties.

In addition to the features proposed in [24], new variables were considered to be useful and were thus added:

- The time between measures: to consider the time gap between the track measures.
- The total time of the segment: as a support variable for the time between measures, considering the total time of the segment, rather than only the measurement time gaps.

Statistical variables, such as the average or the maximum, in addition to ratios, can be useful to summarize information about the overall trajectory. However, additional statistical variables are used to provide more information about the segments; in addition to the mean and the maximum, the proposed approach also uses the mode, minimum, standard deviation, and three quartiles. Finally, variables with additional information, such as ship class, maneuver type, and ship dimensions, are also included. The final feature array is shown in Figure 3. This array provides the classifier with a total of 44 input variables.

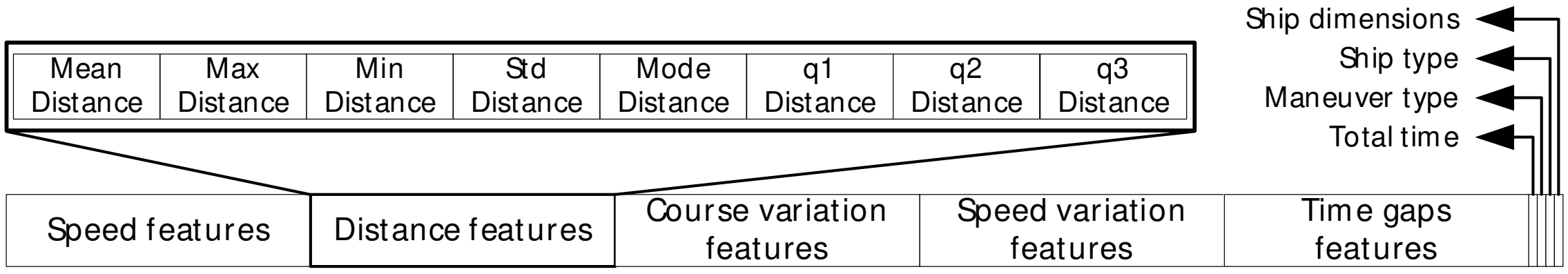

**Figure 3.** Array of kinematic features for classification.

### 3.6.2. Classification Algorithm

The classification is applied to a binary class problem to predict if a tracked ship is a fishing ship. Thus, the algorithms proposed for this approach were selected because of their ability to operate on binary problems, in addition to their demonstrated results and the simple interpretation of their output.

For classification, the training and validation datasets were divided 70:30, ensuring the representation of each ship type by dividing each of the types separately.

The first chosen algorithm is a decision tree that uses the inputs to separate the instances in each tree node. Separations are made using a binary logic rule that divides the instances into two groups according to the variable value. By comparison, the SVM is an algorithm that generates a hyperplane with one dimension for each of the input variables used for the classification, thereby dividing the instances into two different sets, namely, "fishing boats" and "non-fishing boats". These algorithms are among the most widely used in related previous research, with good results shown in trajectory classification problems.

Decision trees also allow analysis of the importance of different input variables in the decision process using the predictor importance function of MATLAB [25]. This makes it possible to evaluate the usefulness of the input variables to the classification and, conversely, to identify those variables that can be ignored to achieve a classification with minimal information.

## 4. Experiments and Results

In this section, the experimentation carried out for the present study is shown. Firstly, an example of the overall process is shown in which each of the phases is explained. Secondly, an explanation of the different experiments carried out is included, as well as the analysis of the results of these experiments. The final section presents a predictor importance analysis for the obtained results.

### *4.1. Ilustrative Example*

To illustrate the overall process, each phase is shown in detail. Starting from the original dataset, the process at each phase is applied, and the result is shown with a summary of the remaining information at each step.

The process began with a dataset spanning three days, in which there were contacts of multiple vessels without any treatment, sorted by date, as shown in Figure 3. The first step was to separate the data according to the MMSI, generating tracks for each of the ships. In Figure 4a, this step is combined with data cleaning, in which trajectories are removed whose information is not sufficient for the following process.

An overview of the data distribution after this cleaning process is shown Figures 4b and 5. As shown, after the process, 22% of the almost 30 million AIS contacts were stored. The main focus of cleaning is on non-consecutive data contacts. This shows that real-world data is not close to the AIS standard, which is possibly a serious problem. With this strict decision, better filtered tracks were available and the final dataset was sufficiently large. Nonetheless, 24% of the original contacts did not have the minimum information to make this classification, indicating that, although AIS is a standard, it contains numerous gaps and inconsistencies.

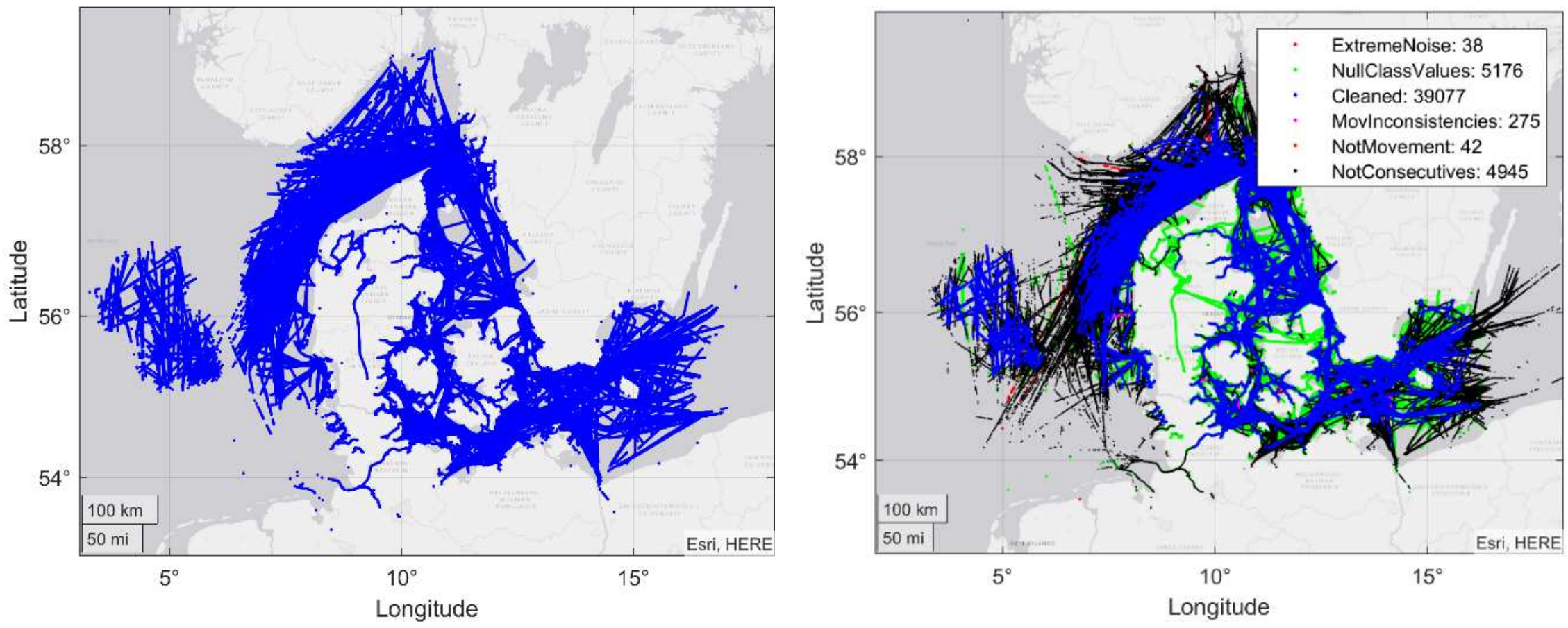

(**a**) (**b**)

**Figure 4.** (**a**) Original AIS contacts on a map; (**b**) AIS trajectories on a map after cleaning.

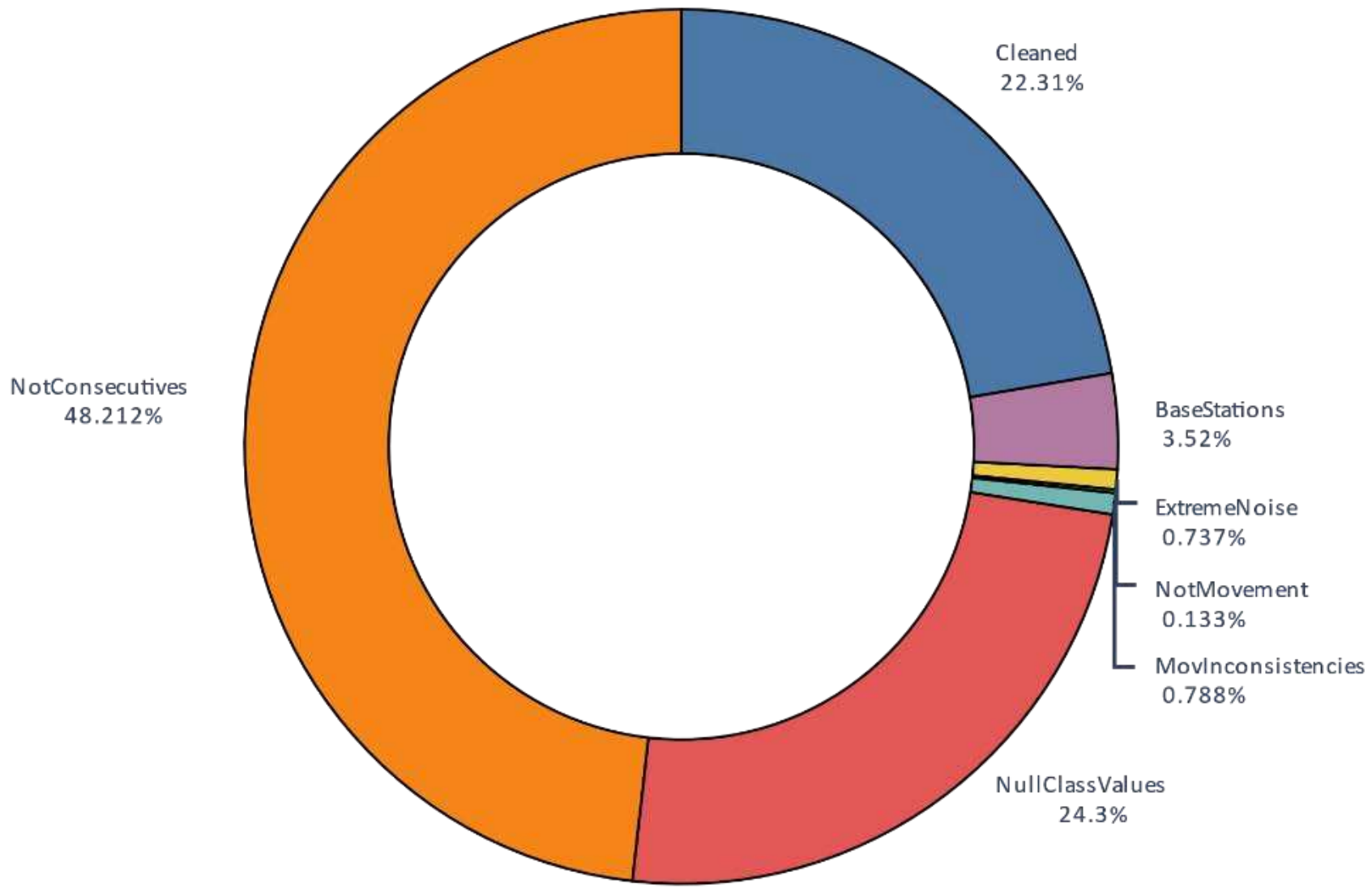

**Figure 5.** Distribution of contacts.

Finally, there were just over 7 million AIS contacts, divided into 39,077 tracks with the desired characteristics. From this set of vessels, the ship with MMSI *205451000* was selected to illustrate the following steps. This ship is a cargo ship and analysis of its data shows that it repeated the same trajectory. During the cleaning process, this vessel's contacts were divided into 18 tracks, as shown in the example in Figure 6a.

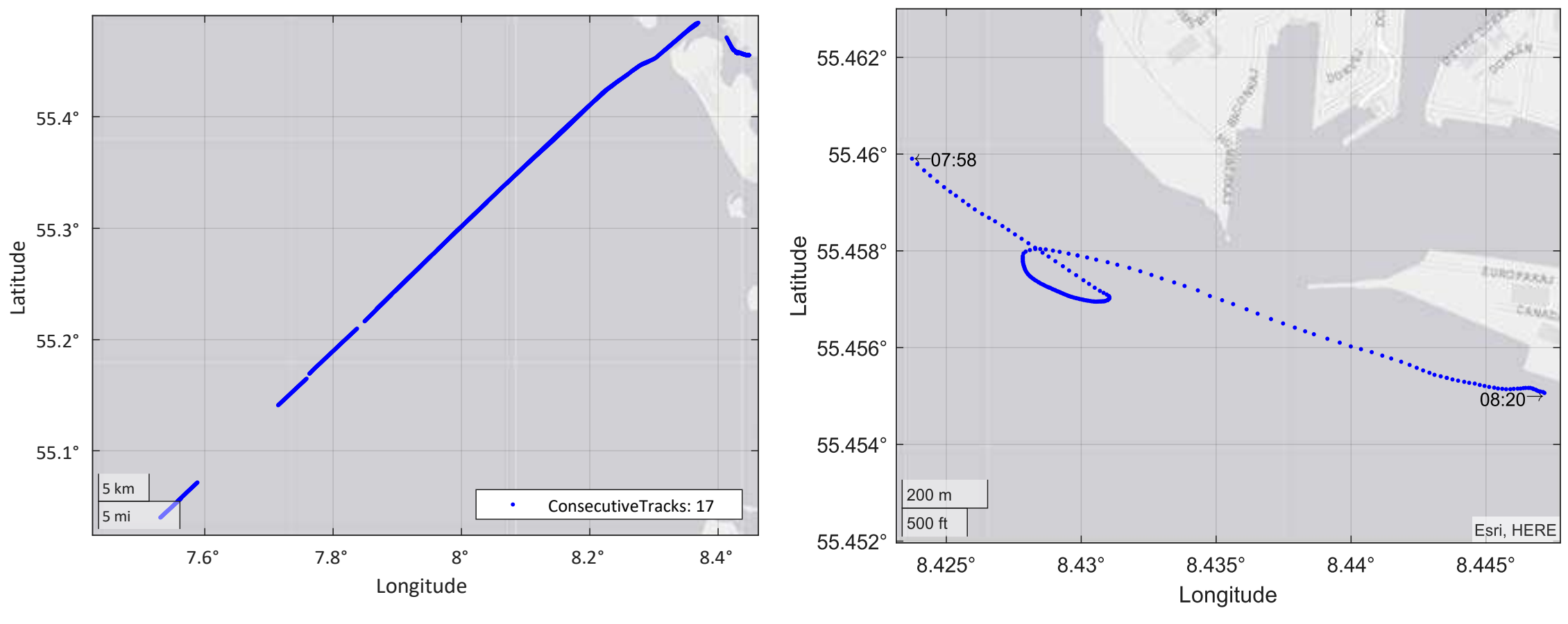

(**a**) (**b**)

**Figure 6.** (**a**) Cleaning process of the example ship; (**b**) Selected track example.

The next step was the filter that reduced the noise of each of the ship tracks. Then, the segmentation divided the trajectories into comparable segments and their statistical information was extracted from each segment for classification.

The image shown in Figure 6b represents an example track of the selected vessel, entering the port, while the graphics shown in Figure 7 represent the different features extracted for that track, in addition to total time. In the figure, the different segments generated are also represented.

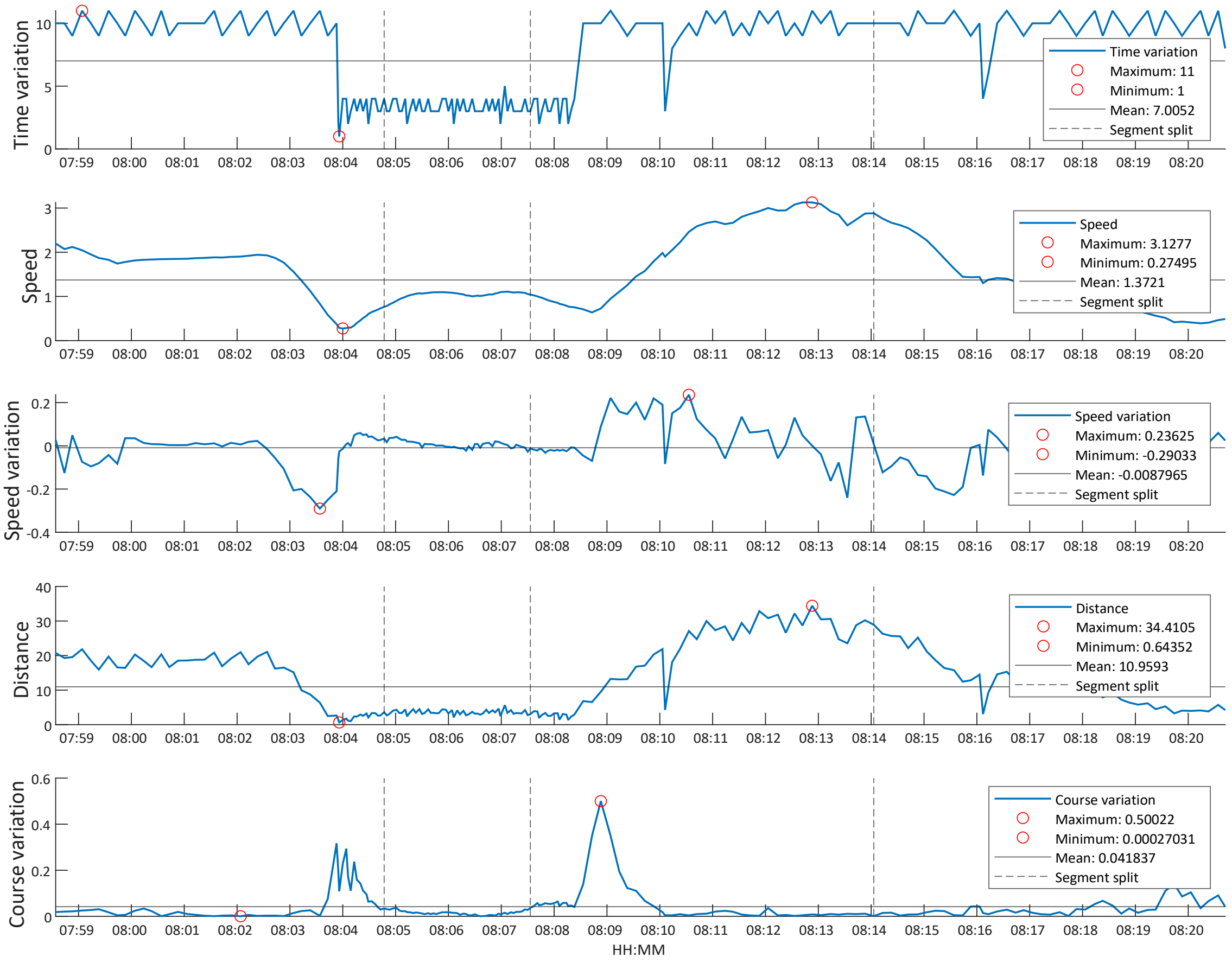

**Figure 7.** Extracted feature example.

Figure 7 shows that the target is affected by speed variations (accelerations and decelerations) when performing a turning maneuver. The figure can also show changes in travel distance or clear changes in direction. Notable, the ship also satisfies the AIS requirements, transmitting more frequently when turning, thus reducing the time intervals.

This track generated four different segments. The feature segments entered the classifier, resulting in a total of 118,283 segments to classify across all of the used instances. To understand the final input data of the classifier, Figure 8 shows the distribution of the values of the class to be classified, i.e., ship type, throughout the process. The "minimal" class was created for the figure by summing the classes with fewer than 25 different MMSIs in each.

Figure 8 shows, in addition to the imbalance problem, the percentage differences of the ship types after cleaning. This is due to many factors: the quality difference of the AIS transponders, breaking of the 11 s time gap, and the trajectory of the vessel itself, which may be stopped for more or less time.

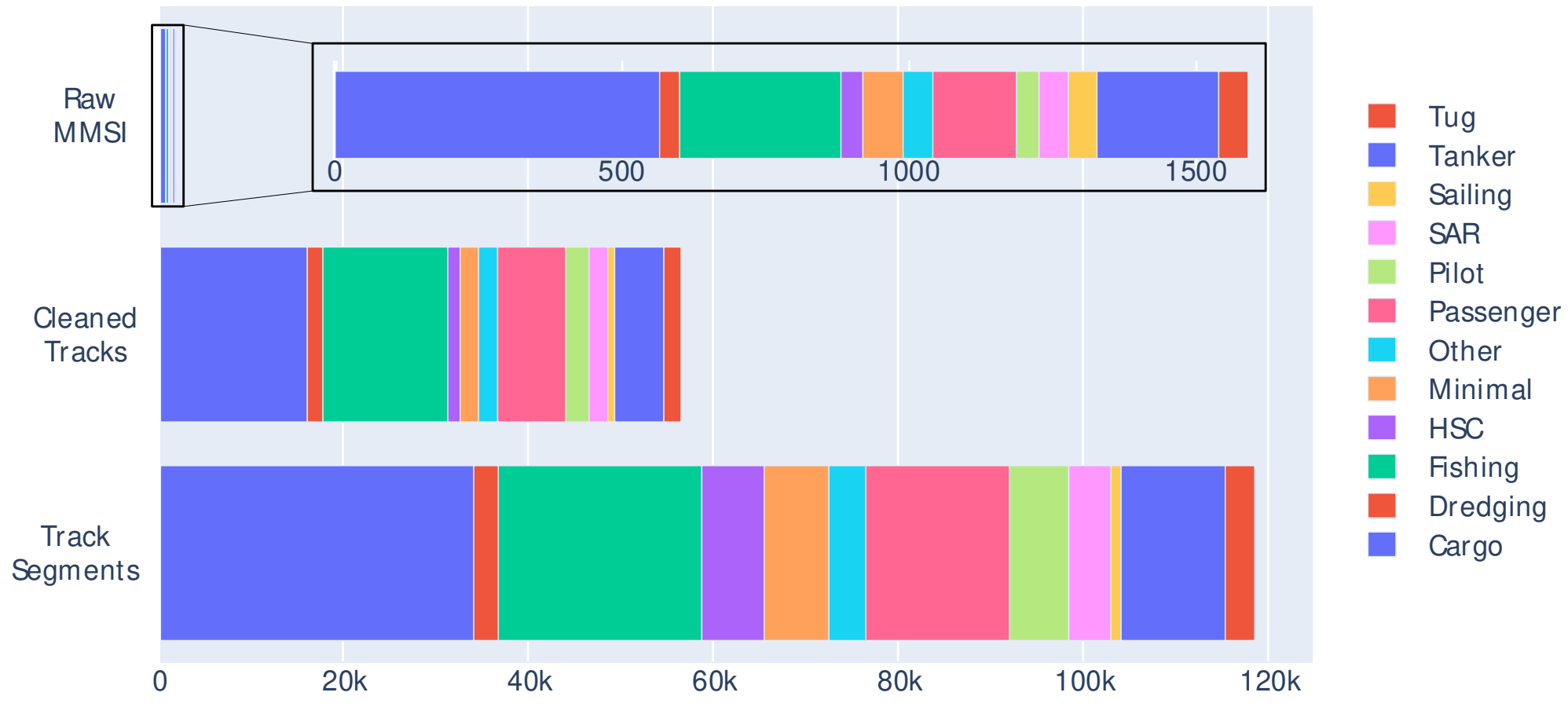

**Figure 8.** Ship type class distribution.

It can also be observed that this segmentation generates more segments if the trajectory is long, which can be a long-term problem if there are ships with many segments. In turn, this may condition the classification of that ship type to the movement of that particular ship, which may not be representative of the class.

## *4.2. Proposed Experimentation*

In this paper, a process of data preparation is proposed with the aim of achieving a ship type classification using only kinematic information about the ship. To test this process, experimentation must evaluate the usefulness of each of the process phases (i.e., cleaning, filtering, imbalance management, track segmentation, and classification) to identify those that contribute a substantial improvement in the classification results. To test the proposed process, experiments were carried out on each of the phases to demonstrate their usefulness.

Data cleaning: Data preparation begins with the cleaning step, which eliminates typical data mining problems. It is not possible to test cleaning using the original dataset directly in the filtering phase, because elements exist within the dataset that obstruct the operation of future algorithms. Therefore, basic cleaning of the essential elements of the process is required. The sub-processes that do not need to be subjected to the general cleaning process are the data limitations, the extreme noise data removal, and the motionless tracks removal. In addition, it is necessary to eliminate null classes and inconsistencies in the data.

Data filtering: The filtering process that reduces noise within the tracks may not be applied when directly passing the intermediate dataset of "Cleaned but noisy tracks" to the data segmentation process.

Data segmentation: The track segmentation may not be applied by entering the IMM filter result into the "preprocessed segments" dataset, passing each entire trajectory as a segment.

Data imbalance problem: One of the main features of this paper is the use of data that represent a clear imbalance between classes, thus prompting the use of rebalancing algorithms to modify the dataset. Notably, many previous studies did not take this problem into consideration or avoided it by using very small datasets.

One of the proposed balancing algorithms (random undersampling or SMOTE) can be applied to the "segment features" dataset. Three datasets can be classified: one in which no balancing algorithm has been applied, and one for each of the algorithms.

From these variations a set of experiments was generated which were grouped into different settings according to the objective; the specific configuration of each of the settings is shown in Table 3, where the symbol X indicates the sub-process was carried out, and the – symbol indicates the sub-process was not carried out. In addition, more tests were performed with the balancing and classification sub-processes because of their importance within the proposed approach. This implies that each type of experiment generated results from all of the possible combinations for the balanced dataset and the classification algorithm; these are marked with the symbol ■ within the table and with the total number of experiments is indicated in the last row.

The "complete process" setting represents the experiments carried out using the proposed process outlined in this article. The objective of these experiments was to test the different configurations of class balancing and the classification of trajectories in the dataset prepared by the proposed process. The remainder of the experiments also tested the class-balancing and classification configurations, although their final objective was to evaluate the different phases of the process. The "no cleaning" setting evaluated the cleaning step in the proposed process, "no filtering" evaluated the use of a filtering algorithm, and the "no segmentation" setting evaluated the track segmentation step.

**Table 3.** Experiments types included subprocess.

| Process | Subprocess | Complete Process | No Cleaning | No Filtering | No Segmentation |
|---|---|---|---|---|---|
| Cleaning | Full cleaning process | X | - | X | X |
| | Only minimum necessary cleaning | - | X | - | - |
| Filtering | Filtered dataset | X | X | - | X |
| | Non filtered dataset | - | - | X | - |
| Segmentation | Segments dataset | X | X | X | - |
| | Full tracks dataset | - | - | - | X |
| Imbalance management | No algorithm | ■ | ■ | ■ | ■ |
| | Random undersampling | ■ | ■ | ■ | ■ |
| | SMOTE | ■ | ■ | ■ | ■ |
| Classification | Decision trees | ■ | ■ | ■ | ■ |
| | SVM | ■ | ■ | ■ | ■ |
| Predictor importance analysis | | Yes | No | No | No |
| Total experiments | | 12 | 6 | 6 | 6 |

The final results of this experiment allowed analysis of the different phases of the proposed process. In addition, within the analysis a further study was carried out on the importance of the classifier inputs; thus, it was possible to study the ability of the extracted features to detect fishing ships using minimal information.

To evaluate the classification problem using balanced data, the accuracy metric and F measure were used. The first metric measured the overall results of the classification, while the second was applied specifically to the minority class to measure the specific results of its classification.

The importance of each variable during classification was measured using the predictor importance calculated in MATLAB with the results of the decision tree classification, thereby obtaining an approximation of the variables that best divide each dataset. This evaluation, as shown in the table, was applied over the complete process setting, and new experiments were added to check that the variables marked as important genuinely provided most of the class division information. Thus, it was necessary to undertake a second evaluation of a classification that uses only the selected input parameters.

### *4.3. Experimentation Results*

Class imbalance generates a multi-objective problem, within which the solutions do not satisfy the accuracy metric and the F-measure simultaneously; one of the metrics is usually favored over the other. This, it is important to note that there is no optimal solution to the problem. Figure 9 shows a diagram in which all of the proposed experiments are plotted using the two metrics as coordinate axes. Thus, several points showing good results can be observed, although none stands out clearly from the rest. Although there is no clear winner, the best solutions represent a compromise between both metrics and were mostly derived from the complete process setting.

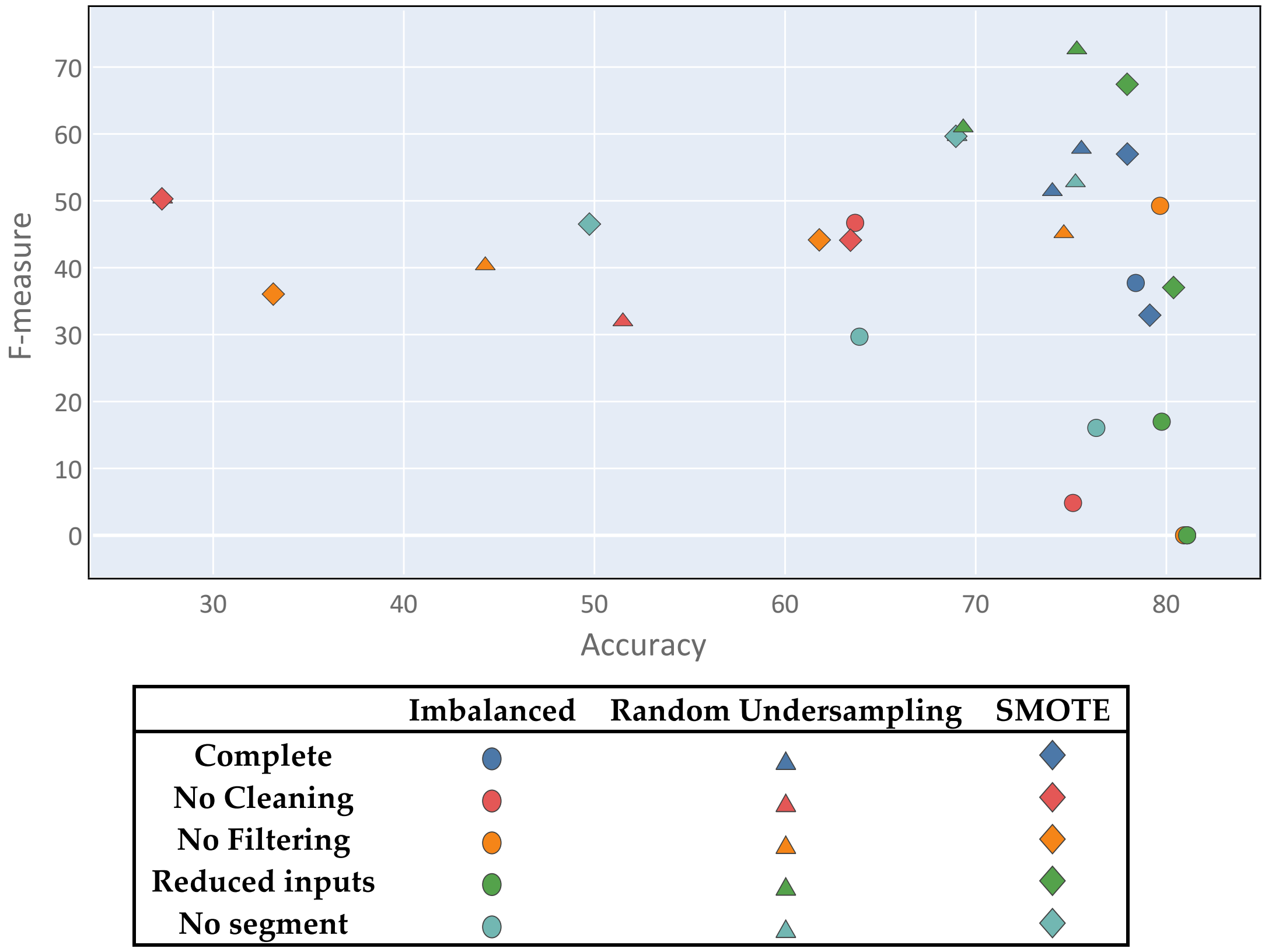

**Figure 9.** Multi-objective problem visualization.

The following figures present the success of the classification approaches with the different tested algorithms. The bars show the metrics used for the performance evaluation, namely, accuracy in blue and F-measure in red, whereas each row shows the approximation to the imbalance problem. The first rows show the results from the SMOTE algorithm, the second show random undersampling results, and third show the results from classification with an unbalanced dataset.

Figure 10a shows the decision tree algorithm results; the accuracy value was decreased with the resampling algorithms, as expected, and an improvement in the F-measure. The F-measure was not high because the decision tree algorithm is not a discriminative algorithm that generates a large number of false positives or false negatives (which are the results that reduce the F-measure by affecting the sensitivity and the precision).

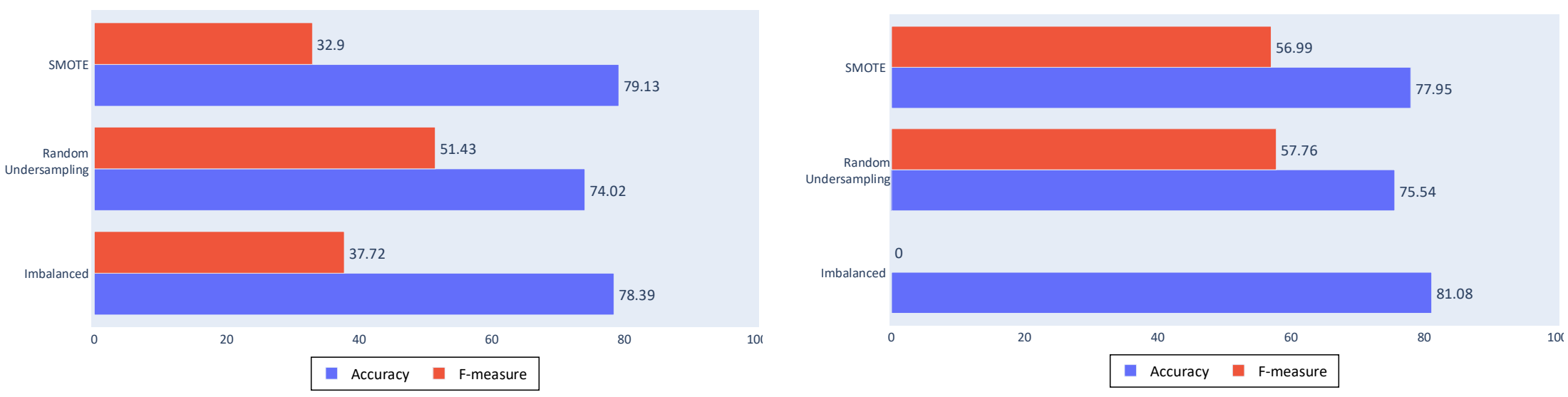

(**a**) (**b**)

**Figure 10.** (**a**) Complete process decision tree results; (**b**) Complete process Support Vector Machine (SVM) results.

Figure 10b shows that the SVM obtained a similar accuracy, although it shows more discrimination with an F-measure of 0 for the imbalanced dataset. This is a poor result for the proposed approach as it means that almost all fishing ships were classified as non-fishing. Although accuracy was slightly lower, the improvement provided by the balancing algorithms can be seen in the F-measure, not only because of the major increase, but also because of the improved values.

Using the classification generated by the decision trees, the importance of the different used predictors can be considered and the importance of each input can be evaluated. Figure 11a shows that the minimum and mode statistics are barely relevant in the classification, whereas the most useful kinematics are the speed, total time, and course variation, as seen in Figure 11b.

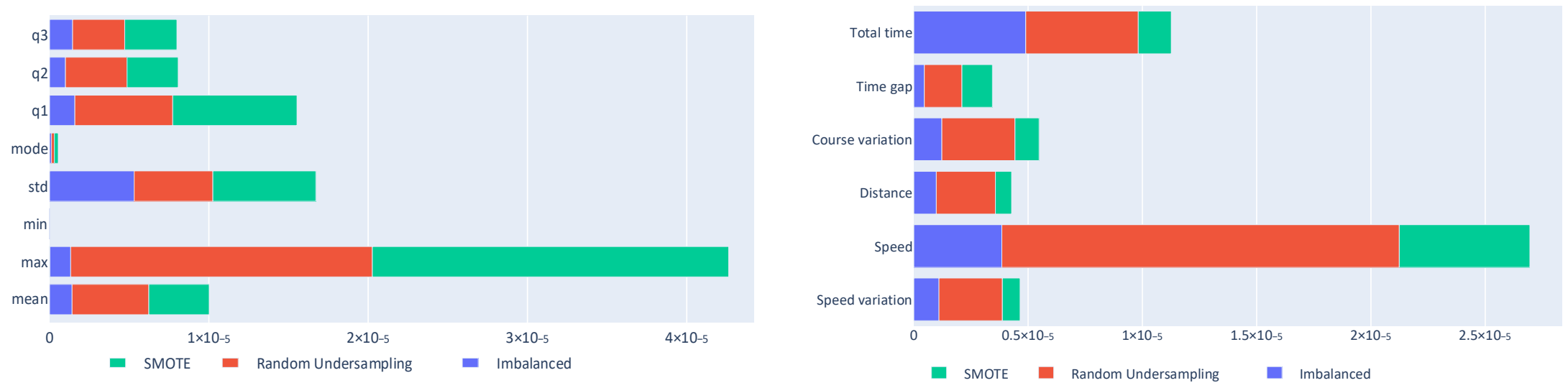

(**a**) (**b**)

**Figure 11.** (**a**) Statistics predictor importance; (**b**) Feature predictor importance.

To test these conclusions, Figure 12a,b show the results of a classification made only with these kinematics and all of the statistics except the minimum and the mode. These reduced inputs results are similar to those of the complete process, which implies that it is possible to achieve fishing ship classification using only three kinematic values, although other kinematics can be useful in different problems.

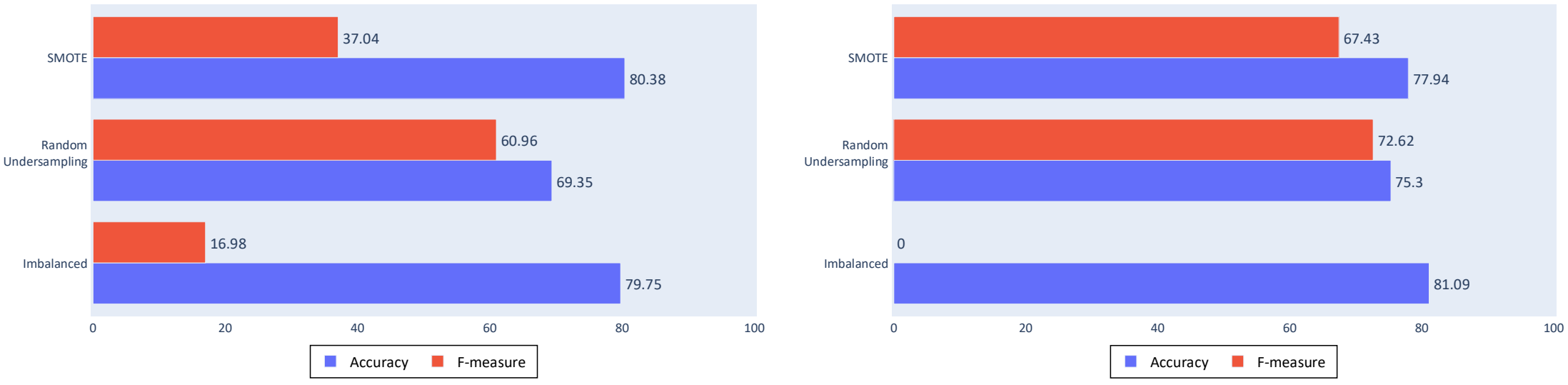

(**a**) (**b**)

**Figure 12.** (**a**) Decision tree results with reduced inputs; (**b**) SVM results with reduced inputs.

Of the three tested settings, the cleaning process shows the greatest variation from the complete process results. This result shows that cleaning is an important process within the proposed approach, because comparing results of a complete cleaning and a minimum cleaning shows a worsening in all of the experiments using the latter, independently of the classifier or the balancing algorithm used.

As seen in Figure 13a,b, accuracy is the most affected measure by the change to a minimum cleaning. The balanced datasets were most affected because the algorithms that generate these sets depend on the quality of the data.

The filtering process in Figure 14a,b shows similar results to those obtained by the complete process for the imbalanced dataset, maintaining the classification made for the majority class. However, for the balanced datasets, a worsening can be noted. This is again due to the quality of data used in the balancing algorithms because the filtered tracks include less noise than the unfiltered tracks.

Non-segmentation in Figure 15a,b generally also yielded similar results to the complete process, although these results were still worse for the less comparable classification inputs. This can be seen particularly in the SVM classifier or in the use of the SMOTE algorithm.

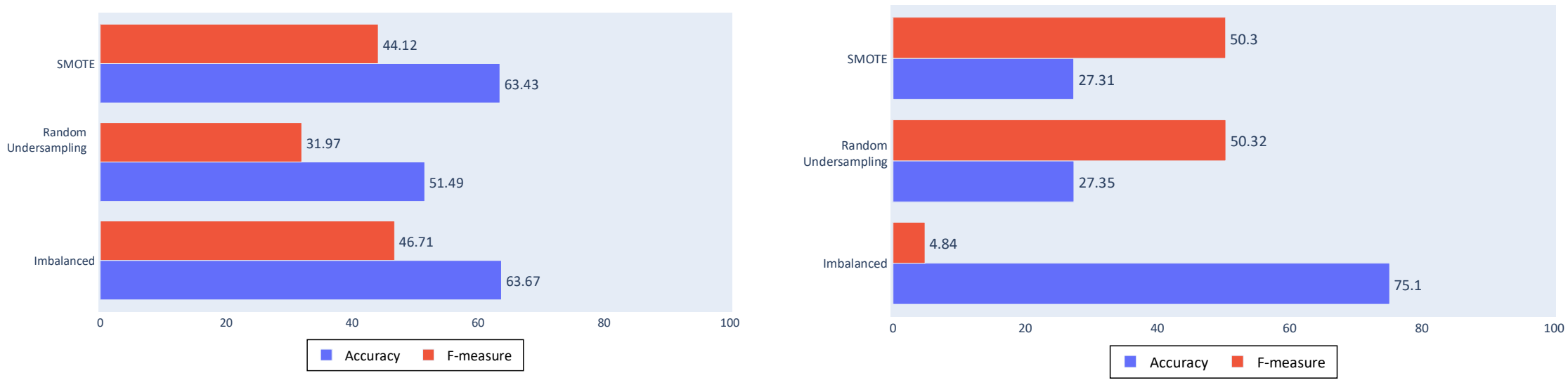

(**a**) (**b**)

**Figure 13.** (**a**) No cleaning decision tree results; (**b**) No cleaning SVM results.

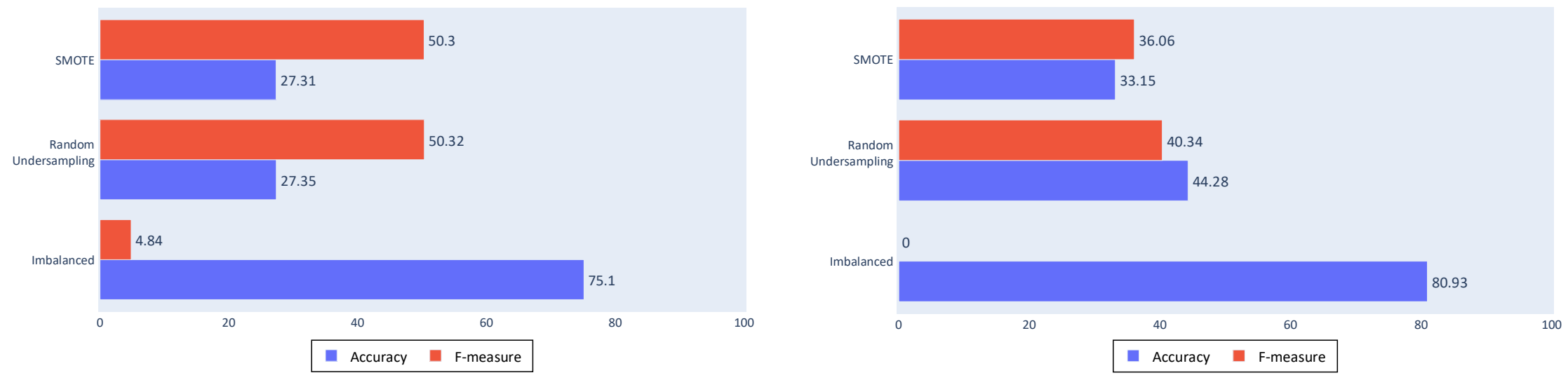

(**a**) (**b**)

**Figure 14.** (**a**) No filtering decision tree results; (**b**) No filtering SVM results.

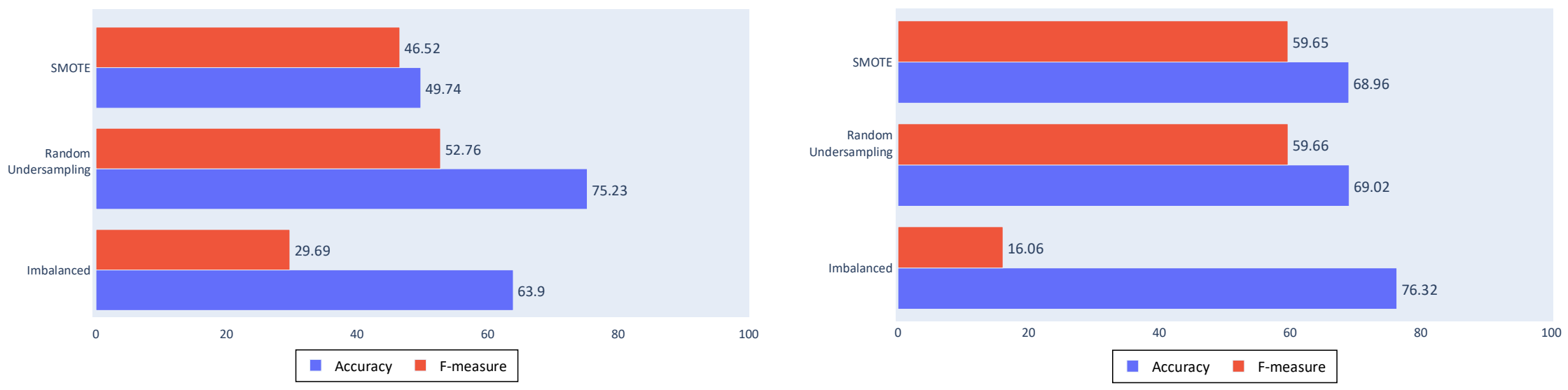

(**a**) (**b**)

**Figure 15.** (**a**) No segmentation decision tree results; (**b**) No segmentation SVM results.

It is important to note that although the three types of experiments show a worsening for the two types of tested classifiers; it can be seen that the SVM results are affected more than those of the decision trees, showing results that are usually worse even though the yielded better results in the complete process experiment.

As a summary of the results obtained from the experimentation, the following conclusions can be obtained:

- With the complete process, the SVM yields better results with the balanced datasets, and resulted in the best results.
- The cleaning step is the most significant phase due to the worsening of results in its non-realization setting.
- Not filtering affects the SVM and balanced sets results in particular.
- Non-segmentation particularly affects the results when SMOTE is used for balancing.
- In all of the settings with an imbalanced dataset, the decision trees algorithm provides better results than the SVM, since its F-measure is less reduced.
- In terms of accuracy, the imbalanced SVM obtains the best results, although, in return, the low values of the F-measure indicate poor fishing ship detection.

### *4.4. K-Fold Validation*

Finally, to validate the accuracy of the obtained results, k-fold cross validation can be used over the best results. Since the problem presents a multiple objective between the F-measure and the accuracy metric, the best results for evaluation were those on the Pareto front highlighted with the blue line in Figure 16.

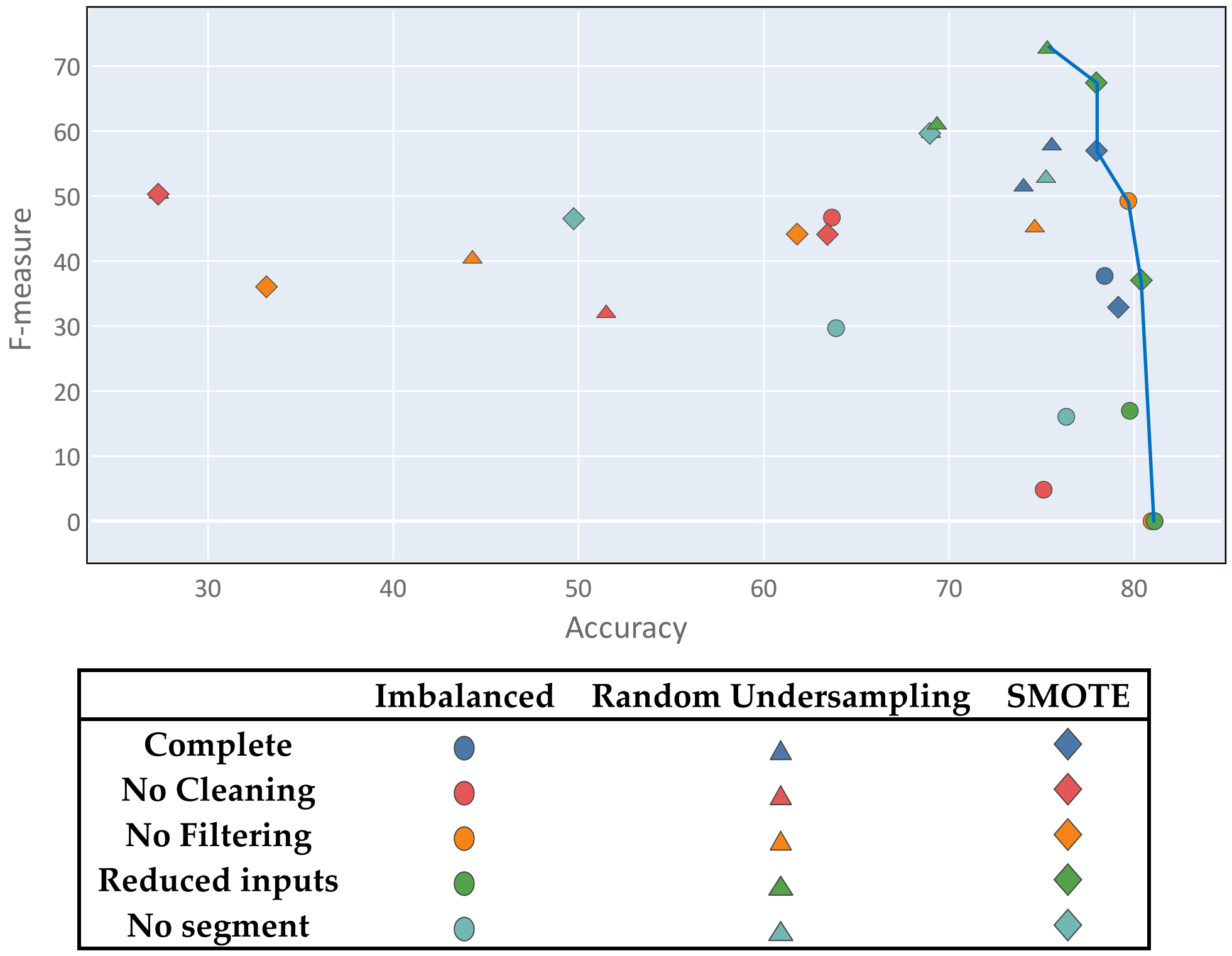

**Figure 16.** Multi-objective problem Pareto front.

For the result to be valid, the balancing algorithm must not be applied to the test data; therefore the k-fold algorithm generated 10 divisions, and iteratively chose nine as the training set on which the balancing was applied. In each iteration, the nine divisions had only original data, which means that no balancing algorithm was applied from previous iterations.

Figure 17 contains a breakdown of the results obtained by the k-fold algorithm. For each experiment the figure shows the 10 individual k-fold runs (green and red thin columns). The average of these runs is the value of the k-fold run, shown as the blue and red background columns). The figure also shows the result of the previous 70—30 experimentation (blue and red diamonds).

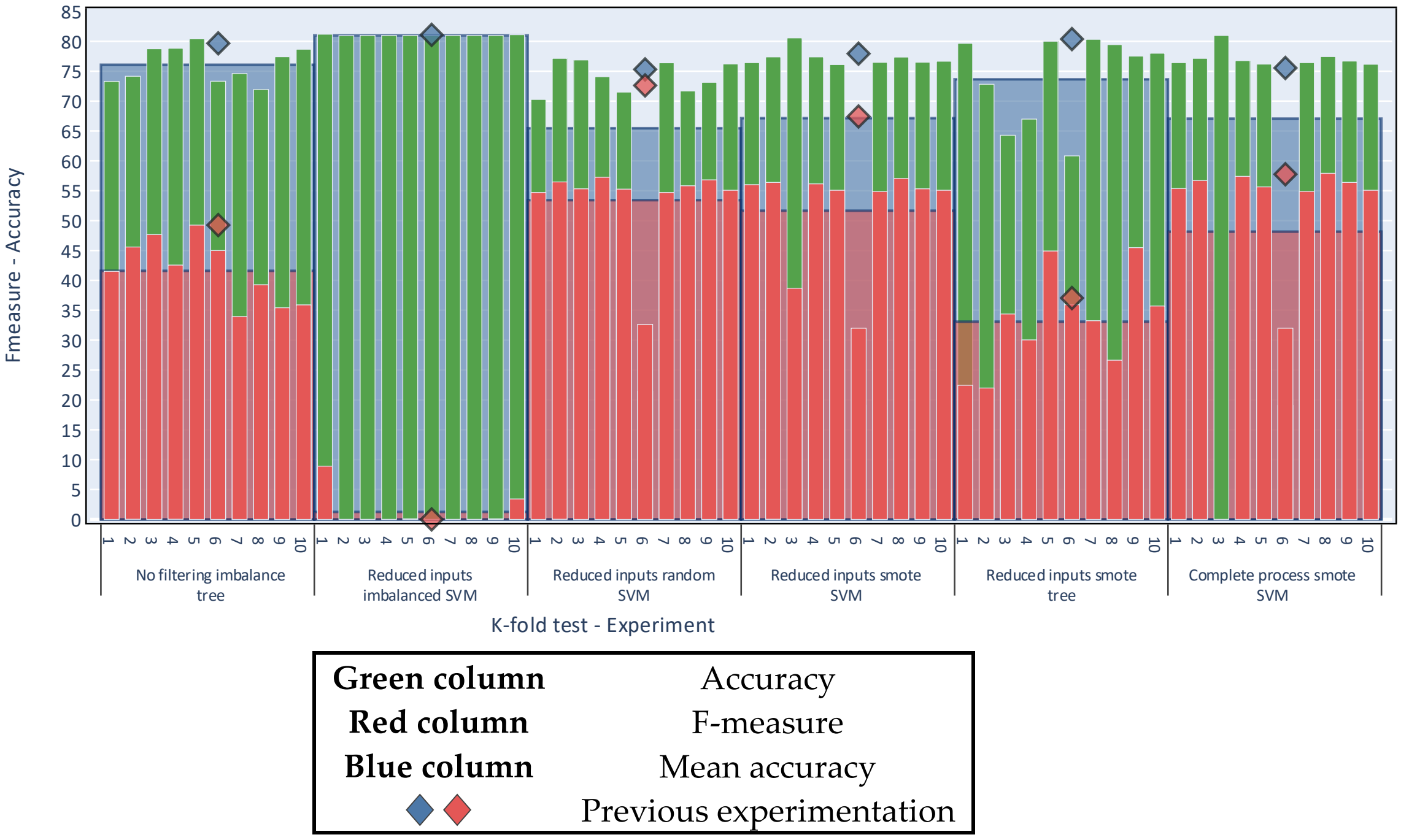

**Figure 17.** K-fold cross validation.

Figure 17 shows that performing the evaluation with a random 70:30 split for training and testing yields results consistent with the evaluation made by K-fold cross validation with k = 10. With this new form of validation, the conclusions remain unchanged. That is, considering only the accuracy metric, the best results are derived for unbalanced sets; by comparison, in also considering the F-measure and following the criteria of the multi-objective problem, balancing acquires greater importance. Finally, the dataset with reduced inputs continues to obtain the best results in both metrics.

## 5. Conclusions and Perspectives

The proposed process demonstrates the ability to prepare data for the desired classification of fishing vessels. This is evident both in the results demonstrated by the complete process experiment and in those of the other experiments.

All of the conducted experiments demonstrate the usefulness of the different steps of the proposed process discussed in this article. However, although the process steps prove to be effective, there is scope for improvement by using more effective algorithms. For example, the segmentation used in the proposed approach simply generates fragments of a fixed size; it is therefore likely that analyzing the same problem with fragments of variable sizes would provide more information. Thus, a possible future improvement is the analysis of alternative algorithms that can provide new approaches to the problem, improve the existing algorithms, or allow variations of algorithms to be tested.

An element to emphasize is that acceptable results are obtained using minimal information. The proposed approach achieves classification results that are close to those from the complete process experiment using only the speed, total time, and course variation kinematics.

The proposed classification approach could also be improved in the future with the addition of features extracted from new sources of information, such as satellites or newly developed reliable sensors. These technologies can provide new information to improve understanding of the context and, as a result, ease the detection of fishing vessels.

**Author Contributions:** Conceptualization, D.A., D.S.P., J.M.M.; Data Curation, D.A., D.S.P.; Formal Analysis, D.A., D.S.P., J.G.; Funding Acquisition, J.G, J.M.M.; Investigation, D.A., D.S.P.; Methodology, D.A., D.S.P.; Project Administration, J.G, J.M.M.; Resources, D.A., D.S.P., J.G, J.M.M.; Software, D.A., D.S.P.; Supervision, J.G, J.M.M.; Validation, D.A., D.S.P., J.G, J.M.M.; Visualization, D.A., D.S.P.; Writing-Original Draft Preparation, D.A., D.S.P.; Writing-Review & Editing, D.A., D.S.P., J.G, J.M.M. All authors have read and agreed to the published version of the manuscript.

**Funding:** This research was funded by public research projects of Spanish Ministry of Economy and Competitivity (MINECO), reference TEC2017-88048-C2-2-R.

**Conflicts of Interest:** The authors declare no conflict of interest.